\documentclass[5pt]{article}

\usepackage[letterpaper]{geometry}
\usepackage{spconf,amsmath,epsfig}
\usepackage{enumitem}

\usepackage{fancyhdr}
\begin{document}\sloppy

\def\x{{\mathbf x}}
\def\L{{\cal L}}

\title{ODN: Opening the Deep Network for Open-set Action Recognition}
%

\name{Yu Shu$^{1}$, Yemin Shi$^{1}$, Yaowei Wang$^{2*}$, Yixiong Zou$^{1}$, Qingsheng Yuan$^{3}$, Yonghong Tian$^{1*}$\thanks{Corresponding author: Yaowei Wang (email: yaoweiwang@bit.edu.cn) and Yonghong Tian (email: yhtian@pku.edu.cn).}}
\address{$^{1}$ National Engineering Laboratory for Video Technology,
	School of EE\&CS, \\
	Peking University, Beijing, China\\
	$^{2}$ School of Information and Electronics, Beijing Institute of Technology, Beijing, China \\
	$^{3}$ School of Cyber Security, University of Chinese Academy of Sciences, Beijing, China
}

%
%
%

\maketitle

\begin{abstract}
In recent years, the performance of action recognition has been significantly improved with the help of deep neural networks. Most of the existing action recognition works hold the \textit{closed-set} assumption that all action categories are known beforehand while deep networks can be well trained for these categories. However, action recognition in the real world is essentially an \textit{open-set} problem, namely, it is impossible to know all action categories beforehand and consequently infeasible to prepare sufficient training samples for those emerging categories. In this case, applying closed-set recognition methods will definitely lead to unseen-category errors. To address this challenge, we propose the Open Deep Network (ODN) for the open-set action recognition task. Technologically, ODN detects new categories by applying a multi-class triplet thresholding method, and then dynamically reconstructs the classification layer and ``opens'' the deep network by adding predictors for new categories continually. In order to transfer the learned knowledge to the new category, two novel methods, Emphasis Initialization and Allometry Training, are adopted to initialize and incrementally train the new predictor so that only few samples are needed to fine-tune the model. Extensive experiments show that ODN can effectively detect and recognize new categories with little human intervention, thus applicable to the open-set action recognition tasks in the real world. Moreover, ODN can even achieve comparable performance to some closed-set methods.
\end{abstract}
\section{Introduction}
\label{sec:intro}

Deep neural networks have made remarkable achievements in many visual recognition problems. Specifically, the performance of human action recognition has been improved significantly in recent works \cite{wang2016temporal,shi2017learning}. However, most of these recognition systems are designed for a static closed world, where the primary assumption is that all categories are known as priori. However, action recognition in the real world is essentially an \textit{open-set} problem, namely, it is impossible to know all action categories beforehand and consequently infeasible to prepare sufficient training samples to train the deep networks for those emerging categories. In this case, applying closed-set recognition methods will definitely lead to unseen-category errors. Therefore, an open-set method is more desirable for the real-wolrd action recognition applications.

\begin{figure}
	\centerline{\includegraphics[width=0.40\textwidth]{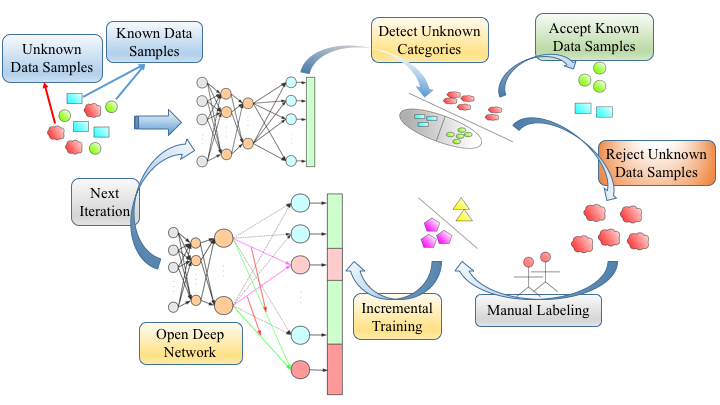}}
	\caption{The workflow of ODN. ODN first trains an initial model on the knowns. In the open-set scenario, ODN detects the unknowns. Then a strong teacher is invoked to label the unknowns. Next, ODN dynamically incorporates the new category in the model. Finally, ODN fine-tunes the model with only few samples to make the unknowns "known".}
	\label{figure:png}
\end{figure}

Technologically speaking, many methods on incremental learning can be used to handle new instances of known categories \cite{Tveit2003Multicategory,yeh2008dynamic}. However, most of these approaches cannot handle unknowns or dynamically add new categories to the system. In \cite{mensink2012metric}, new categories are added according to the mean features. This approach, however, assumes that the number of initial categories is relatively large. An alternative multi-class incremental approach based on least-squares SVM has been proposed by Kuzborskij \textit{et al.} \cite{kuzborskij2013n}. Each time a category is added, the whole set of the hyperplanes will be updated, which is too expensive as the number of categories grows.  

In particular, researches on open set recognition \cite{scheirer2013toward,scheirer2014probability,bendale2016towards}  have established formulations of rejecting unknown objects. Among them, Abhijit \textit{et al.} \cite{bendale2016towards} adapted the concept of Meta-Recognition \cite{scheirer2011meta,zhang2014predicting} to deep networks, and proposed an OpenMax method. However, the issue of incremental learning with additional category remains an unsolved problem. Meanwhile, Abhijit \textit{et al.} \cite{bendale2015towards} also proposed a SVM-based recognition system that could continuously learn new object categories by extending the NCM-like algorithms \cite{Ristin2014Incremental} to a NNO algorithm, but it is not applicable in deep networks. Therefore, the open-set recognition with deep networks is still a challenging problem and needs more efforts to be devoted.

In this study, we extend the open world recognition problem in \cite{bendale2015towards} to video action recognition task. Let the knowns labeled by positive integers $\mathcal{N}^{+}$ and the zero label (0) label the unknowns. And let $\mathcal{K}_{t}^{k} \subset \mathcal{N}^{+}$ be the set of knowns which has \textit{k} categories at time $t$. Let the known category number be \textit{N} in the initial. Thus a typical network for the open-set action recognition problem should include the following four steps:
\begin{enumerate}
	\item \textbf{Multi-class unknown category detection:} A detector $\gamma(\varphi)$ : $\mathcal{R}^{d} \mapsto [0, 1]$ is needed to determine whether the current data sample $\varphi$, which corresponds to features $x\in \mathcal{R}^{d}$, is from an unknown (0) category or not.
	\item \textbf{Unknown category labeling:} $\mathcal{R}^{d} \mapsto \mathcal{N}^{+}$. A strong teacher (human in general) is invoked to identify the unknown categories, which are detected in the previous process, by assigning a new label. Assuming the detection process finding \textit{m} new categories up to time \textit{t}, the set of known categories becomes $\mathcal{K}_{t+1} = \mathcal{K}_{t} \cup \{N+1, N+2, \dots, N+m\} $.
	\item \textbf{Open Deep Network updating:} Recognition process : $\mathcal{P}(\varphi)$: $\mathcal{R}^{d} \mapsto \mathcal{K}^{k}$ at time $t$. Then the network is updated by adding one category in the classification layer at time $t+1$. In this case, the recognition process is expressed as $\mathcal{R}^{d} \mapsto \mathcal{K}^{k+1}$.
	\item \textbf{Incremental training:} The network with $\mathcal{P}(\varphi)$ : $\mathcal{R}^{d} \mapsto \mathcal{K}^{k+1}$ is incrementally fine-tuned with few labeled samples for the new category. 
\end{enumerate}
We can see that in the open-set action recognition problem, the major challenges lie in discovering unknown categories and dynamically opening and updating the deep network.

To address these challenges, this paper proposes the Open Deep Network (ODN) for open-set action recognition with deep networks. Fig. \ref{figure:png} illustrates the workflow of ODN which gradually makes each "unknown" to "known". Extensive experiments on the HMDB51, UCF50 and UCF101 benchmark datasets show that ODN can effectively detect and recognize new categories with little human intervention. 

The main contributions are summarized as follows:
\begin{enumerate}
	\item A network dynamical reconstruction method (ODN) is proposed to support the task of open-set recognition. The novel architecture, ODN, applies a multi-class triplet thresholding method to detect unknown categories and simultaneously transfers the knowledge from known categories to new categories to quickly make "unknown" to "known".
	\item An evaluation protocol is developed for the open-set action recognition task. The experiment results show that ODN is able to achieve comparable performance to the closed-set setting while few new samples are needed.
\end{enumerate}

The rest of this paper is organized as follows: Section \ref{relatedworks} simply reviews the related works on incremental learning, open-set learning and action recognition. Section \ref{Open_Deep_Networks} presents our ODN and Section \ref{experiments} describes its implementation details and experiments. Finally, Section \ref{conclusion} concludes this paper.

\section{Related works}\label{relatedworks}
\noindent \textbf{Incremental learning.} Many incremental methods based on SVM were proposed in these years. Cauwenberghs \textit{et al.} \cite{cauwenberghs2001incremental} proposed an incremental binary SVM by means of saving and updating KKT conditions. Yeh \textit{et al.} \cite{yeh2008dynamic} extended the incremental binary SVM to object recognition and demonstrated multi-class incremental learning. Pronobis \textit{et al.} \cite{pronobis2010more} proposed a memory-controlled online incremental SVM which combined an approximate technique \cite{Syed1999Incremental}. Some other multi-class incremental learning methods \cite{wang2010multi} are trying to increase the number of training samples instead of categories.
\\ 
\noindent \textbf{Open-set learning.} Open-set recognition assumes there are both known and unknown data in the test phase, considering that the knowledge of categories in the real world is not available. Recent works on open-set recognition \cite{scheirer2013toward,scheirer2014probability,bendale2016towards,li2005open} formalized processes of rejecting unknown objects in open-set recognition. In \cite{bendale2015towards},  Abhijit \textit{et al.} proposed a recognition system that suits the open world scenario but is not applicable in deep networks.
\\ 
\noindent \textbf{Action recognition with deep networks.} In recent years, CNNs have achieved state-of-the-art performance on various tasks (e.g. \cite{szegedy2015going,he2015delving}) and it has been proven that features learned from CNNs are much better than the hand-crafted features \cite{dalal2005histograms}. Many works \cite{shi2015learning,zha2015exploiting} have been done to transfer CNNs to video tasks and have made significant progress. The two-stream network \cite{simonyan2014two} is the most widely used baseline video classification model, which has both spatial and temporal networks, ultimately achieves the optimal performance. The latest works \cite{wang2016temporal,shi2017learning} have achieved fantastic performance. However, these works are designed for the closed-set scenario, and how to open the deep networks and dynamically add new categories still remain unsolved.

\section{Open Deep Network}\label{Open_Deep_Networks}
In this section, we will describe the Open Deep Network (ODN) for open-set recognition. The methodology is composed of four parts, which will be elaborated in the following subsections. 

\subsection{Multi-class unknown category detection}
The first step of ODN is to detect the unknowns, considering as rejecting unknowns during the open-set recognition. Scheirer \textit{et al.}\cite{scheirer2011meta} defined the open space risk. When adopting deep networks to handle the open-set recognition, it is necessary to make sure that the open space risk is minimized. But it only gives a general definition and does not specify the method of measuring the distance or which space to measure in. In this paper, we measure the distance in the classification layer of the network. And in the appendix, we proof that ODN can manage the open space risk. 

A natural approach for detecting unknown categories is to apply a threshold on the SoftMax's output. However, we found it not effective in the exploration experiment. Because different categories will have different activation range, a single threshold is not enough for handling these differences. Human action is a composite concept and is the composition of a series of atomic behavioral characteristics. Different human actions may contain several same atomic behavioral characteristics, and these relations may still retain in the activation level. Digging the connected information among the activation values can helps to detect unknown categories. 

Based on above discussion, we propose a multi-class triplet thresholding method combining the inter-class association for detecting unknown categories, which captures the way in which different categories vary in the activation level.

We calculate a triplet threshold $[ \eta, \mu, \delta ]$ per category, i.e. accept threshold $\eta$, reject threshold $\mu$ and distance-reject threshold $\delta$. Define $x_{i,j} \in \mathcal{X}$, where $\mathcal{X}$ = \{$x_{i,j} | x_{i,j}$ is the $j^{th}$ correctly classified sample of category i\} and the total samples number of $\mathcal{X}$ is $X$. Let $F_{i,j} = v_{i}(x_{i,j})$ , $S_{i,j} = \mathbf{Second}(\vec{v}(x_{i,j}))$, where v represents the activation vector of the data sample and the function \textbf{Second}($\vec{v}(x)$) returns the second maximal confidence value of the input vector $\vec{v}(x)$, $\varepsilon$ and $\rho$ are empirical parameters. Then the triplet threshold $[ \eta_{i}, \mu_{i}, \delta_{i} ]$ is calculated as 
\begin{align}
\eta_{i} =& \frac{1}{X}\sum_{j=1}^{X}F_{i,j}\\
\mu_{i} =& \varepsilon * \eta_{i}\\
\delta_{i} =& \rho * \frac{1}{X}\sum_{j=1}^{X}(F_{i,j} - S_{i,j})
\end{align}
During the detecting phase, we accept a data sample as category label \textit{l} only if the index of its top confidence value is \textit{l} and the value is greater than $\eta_{l}$. We consider a sample unknown when all of its confidence value is below $\mu$. And the statistical properties of $\delta$ include correlation information between the two categories, which is a simple way of using the inter-class relation information in the activation level. We use $\delta$ to help detect unknown category in hard samples, which lie between $\eta$ and $\mu$. If the distance is large enough, then we accept the data sample as category label \textit{l}.

\subsection{Unknown category labeling}
A labeling process is applied to the unknown samples detected in the former process. We propose several methods to transfer the learned knowledge to the new categories, hence, only a few new samples will be needed. For some difficult unknown categories, detection process can be repeated to detect more training samples. On average, we only need several manually annotations.

\subsection{Open Deep Network updating}
The open-set recognition requires to update the system to add these new categories found in the multi-class unknown categories detection. One can always retrain the entire system using the known data and new samples, while this solution not only consumes a lot of time and computational resources, but also is easy to be over-fitting because of lacking enough training samples. We propose an updating method by transferring knowledge from the trained model. Our approach helps speed up the training stage and need very few manually annotations.
\[ 
\begin{bmatrix}w_{1}&\mathbf{w_{2}}&w_{3}&\dots&\mathbf{w_{t}}&\dots&w_{N}\end{bmatrix} \quad\qquad\qquad (I)
\]
\[
\begin{bmatrix}w_{1}&\mathbf{w_{2}}&w_{3}&\dots&\mathbf{w_{t}}&\dots&w_{N}&w_{N+1}\end{bmatrix} \qquad (II)
\]
As shown in (I), the model at time \textit{t} has \textit{N} categories corresponding to \textit{N} columns in the transfer weight. The weight contains the knowledge learned in training phase. To open the deep network, we need to add a new column corresponding to the unknown category which we choose to add at this iteration. As show in (II) above, in order to transfer knowledge from the current model, we initialize the $w'_{N+1}:$
\begin{align}
w'_{N+1} = \frac{1}{N}\sum_{n=1}^{N}w_{n}
\end{align}
where $w_{n}$ is the weight column of the $n^{th}$ category.

Considering that human action is a composite concept, the actions are interrelated. Though the action itself is different, some of the action elements which constitute the action may be the same. So the similar categories should play a more critical role in the initialization of $w_{N+1}$. We choose the weight columns of $M$ highest activation values as $w_{m}$ to form the set $\mathcal{H}$. And we define:
\begin{align}
w''_{N+1} = \frac{1}{M}\sum_{m=1}^{M}w_{m}
\end{align}
The $w''_{N+1}$ emphases the similar known categories. Though the unknown category has low activation value for each known category, the weight columns corresponding to higher activation value should be emphasized in initialization. Finally, we blend the overall knowledge and similar emphasis from the current model as:
\begin{align}
w_{N+1} = \alpha \frac{1}{N}\sum_{n=1}^{N}w_{n} + \beta \frac{1}{M}\sum_{m=1}^{M}w_{m}
\end{align}
The $w_{N+1}$ is then concatenated to the transfer weight $W$ to support the new category. And we call this initialization method \textbf{Emphasis Initialization}.

\subsection{Incremental training}
After opening the deep networks, ODN fine-tunes the model using both known and new category samples. As knowledge is transfered from the trained model, it's still feasible in large scale application. To reduce the influence on accuracies of knowns, we adopt a balanced training strategy. We guarantee that each of the known categories has the same number of samples as the new category for fine-tuning, and the known samples can be get from the training set. Because the number of new category samples is small, the entire fine-tuning process finishes quickly.

As for the knowns, the system has been trained before. It is more efficient to train the model at different rates for known and new categories. ODN adopts a different learning rate decay matrix, which varies between known categories and new categories to force new categories learn much faster than known categories during the fine-tuning. We call this method as \textbf{Allometry Training}. It is quite helpful to reduce performance jitter on accuracies of knowns recognition. We define allometry factor $\alpha$ as:
\begin{align}
\alpha_{i} = \left\{
\begin{aligned}
&0.1, ~~~i <= N \\
&~~~1, ~~~i~ >~~ N
\end{aligned}
\right.
\end{align}
When the modified weight transfer matrix is updated with the back propagation algorithm, we implement the allometry factor as learning rate matrix. 

The Allometry Training is inspired by the fine-tuning strategy of CNNs. The main difference is that Allometry Training applies different learning rate for different elements in the same matrix.

\section{Experiments}\label{experiments}
This section will first introduce the details of datasets. Then, we describe the implementation details. Finally, we report the experimental results and give the analysis of results.

\begin{figure*}
	\centerline{\includegraphics[width=0.9\textwidth]{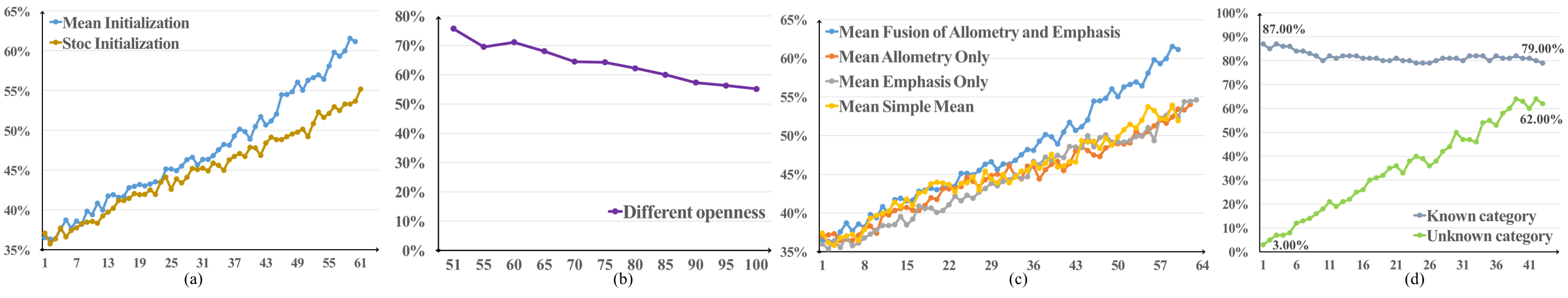}}
	\caption{(a)The comparison between Mean Initialization and Stochastic Initialization during the updating phase. (b)The accuracy of both known and unknown samples with different openness. (c)The illustration of testing accuracy on both known and unknown categories at each iteration. (d)The comparison between the accuracies of known categories and unknown categories at each iteration of the incremental training phase.}
	\label{figure:exploration_experiments}
\end{figure*}

\subsection{Datasets}
To verify the effectiveness of our methods, we conduct experiments on three benchmark datasets: HMDB51 \cite{kuehne2011hmdb}, UCF50 \cite{Kishore2012ucf50} and UCF101 \cite{soomro2012ucf101}.

The HMDB51 dataset contains 6849 clips divided into 51 action categories, each containing a minimum of 101 clips. The UCF50 dataset is an action recognition dataset with 50 action categories, consisting of realistic videos taken from youtube. The UCF101 dataset is one of the most popular action recognition benchmarks. It contains 13,320 video clips from 101 action categories and there are at least 100 video clips for each category. We conduct experiments on the first split of HMDB51 and UCF101.

\subsection{Experiments setting}
\textbf{Training Phase:} The training procedure of ODN has two phases. First, in the initial training phase, known data samples are provided for the CNN to train a classifier with the knowns. Once the initial training phase is completed, the system will use the trained model to get activation values for detecting unknown-category sample process. Second, in the incremental training phase, labeled new category is added to the system one-by-one. And to balance the incremental training phase, we ensure that the sample number of each known category is the same as that of the unknown category.

\noindent\textbf{Testing Phase:} To ensure proper evaluation of ODN, we test it on a set of known and unknown samples which none of them appear in the training phase.

Also, to compare with the traditional closed-set action recognition work, we carried out the experiments on HMDB51 split1 and UCF101 split1, which means we will use much fewer training samples than any other closed-set methods while testing on standard test set.  

\subsection{Implementation details}
To focus on Open Deep Network and simplify the training process, we use a very simple CNN implementation, GoogLeNet \cite{szegedy2015going}, to give a set of exploration experiments. Finally, we give the results on Inception-resnet-v2. In multi-class open set recognition process, we set some empirical parameters. And in our method \textbf{Emphasis Initializion}, we choose 5 corresponding weight columns of similar known categories to emphasize the initiation of the unknown category weight. The details are described in the appendix.

\subsection{Evaluation protocol}
\noindent \textbf{Testing on initial model:} After the initial training phase, we test the initail model on the known samples. And the accuracy on unknown test samples should be 0. 

\noindent \textbf{Testing during the incremental training phase:} After each iteration of the incremental training phase, we test on the entire set of both known and unknown categories. To make the evaluation more rigorous, we treat all samples in the unknown categories by this stage as misidentified. 

\noindent \textbf{Testing phase for the whole system:} Finally, we give the overall performance of the model on the entire test set of both known and unknown categories, and report the accuracy of each category.

\begin{figure}
	\centerline{\includegraphics[width=0.30\textwidth]{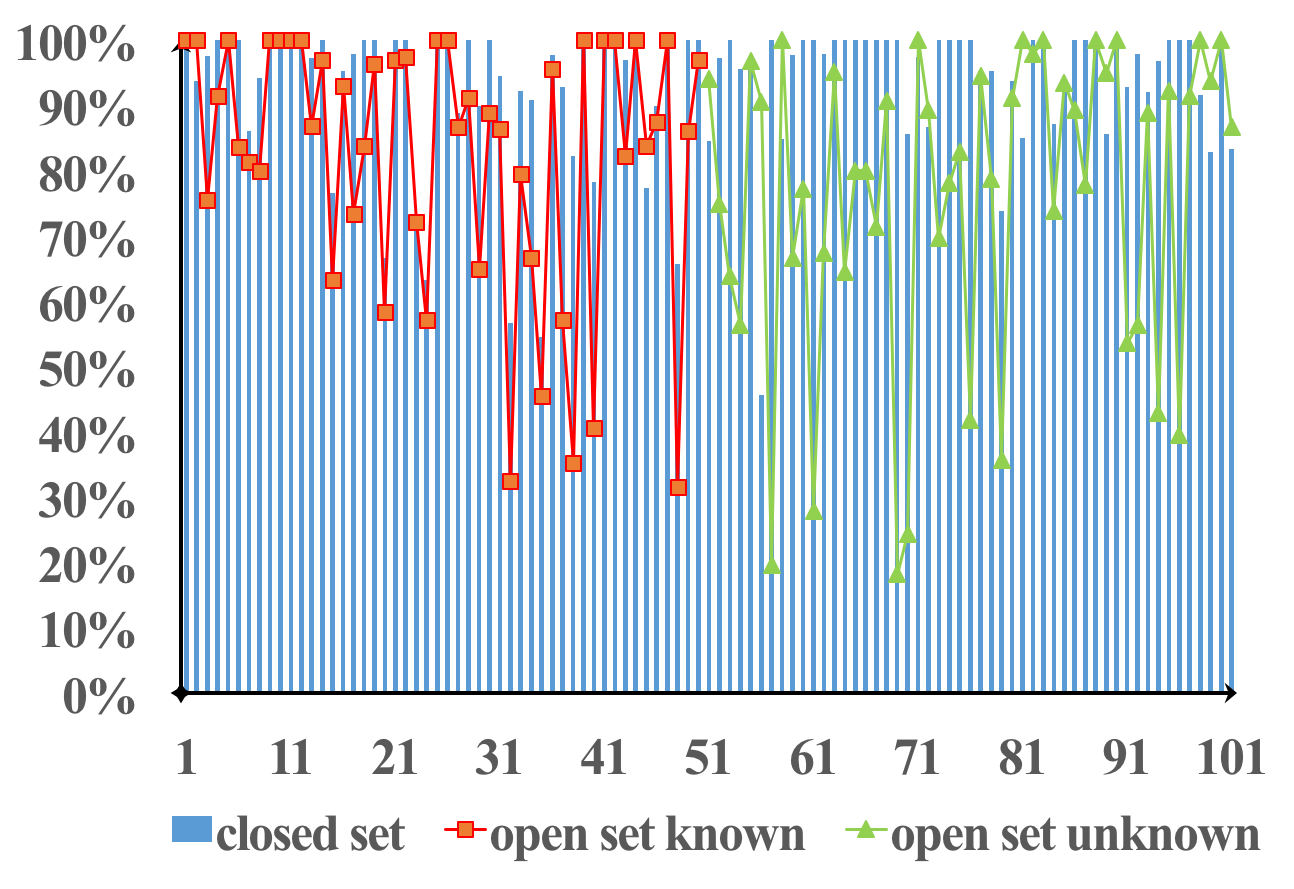}}
	\caption{The comparison between the closed-set and open-set recognition for each category.}
	\label{figure:res_each}
\end{figure}

\subsection{Results and Analysis}
In this section, we will report the results on three datasets, HMDB51, UCF50 and UCF101, and give a detailed analysis. 

We carry out a set of exploration experiments on dataset UCF101, as Fig.\ref{figure:exploration_experiments}, to prove the effectiveness of ODN. We compare the mean initialization method and stochastic initialization method while initializing the new transfer weight. As Fig.\ref{figure:exploration_experiments}(a), we can see the mean initialization method is better for almost all the time, especially when the categories grow larger, for the stochastic initialization would drop lots of knowledge which the system has learned. In the rest of this paper, we will use mean initialization. Fig.\ref{figure:exploration_experiments}(b) shows the system's performance of different openness. We keep the initial model number be 50, half of the category number of UCF101, then let the open world have 1, 5, 10 unknown categories and so on. Along with the unknown categories grow larger, the openness is larger, which means the task is harder. In our evaluation protocol, we should choose half of the dataset categories as known and the other half as unknown. Because this is the most fair setting to evaluation open set models, reducing unknown categories will make the task easier, and increasing lots of unknown categories, one can easily reject most of the input samples as unknown to achieve better performance. 

In Fig.\ref{figure:exploration_experiments}(c) we verify our Emphasis Initialization method and Allometry Training method separately, Emphasis only and Allometry only each improve the performance of the base method Simple mean. In Fig.\ref{figure:exploration_experiments}(d), we show the accuracy of known categories and unknown categories separately. The accuracy of known is decreased, for adding new categories increases the jitter of model. But the accuracy of unknown categories achieves great improvement as the model knowing more categories. 

We also compare with the closed-set recognition. A detailed comparison between the closed-set recognition and ODN is showed in Fig.\ref{figure:res_each}, we test the accuracy of each category both on the closed set and open set. It's amazing that ODN is able to achieve comparable performance with the closed-set recognition, while we only need a teacher to label 5.39 samples on average for each unknown categories, which are detect by ODN. It is much less than averagely 94.4 samples in the closed set recognition. The details of calculating the number of labeling samples are described in the appendix.

\subsection{Comparison with the state-of-the-art methods}
To prove the effectiveness of ODN, we also compare it with several closed-set state-of-the-art action recognition methods in Table\ref{table:compare_state_of_the_art}. Even though using much less training data, our ODN is still able to outperform these closed-set methods.

\begin{table}
	\centering
	\caption{Comparison of ODN to the state-of-the-art closed-set methods. Half of the training set is used to train ODN while full training set is used to train other models.}
	\begin{tabular}{c|c|c|c|c}
		\hline\hline
		Model & UCF50 & Model & HMDB51 & UCF101 \\
		\hline
		Var0\cite{todorovic2012human} & 81.03\% & STIP\cite{kuehne2011hmdb} & 23.0\% & 43.9\% \\
		bIDT\cite{shi2015learning} & 88.57\% & CNN\cite{sadanand2012action} & 26.9\% & 63.3\% \\
		iDT\cite{wang2013action} & 91.2\% & RGB\cite{simonyan2014two} & 40.5\% & 73.0\% \\
		\hline
		ODN & \textbf{93.73\%} & ODN & \textbf{46.01\%} & \textbf{76.07\%} \\
		\hline\hline
	\end{tabular}
	\label{table:compare_state_of_the_art}
\end{table}

\section{Conclusion}\label{conclusion}
In this paper, we propose an Open Deep Network (ODN) for open-set action recognition. In ODN, we propose a multi-class triplet thresholding method for unknown category detection, and Emphasis Initialization and Allometry Training methods to transfer the learned knowledge from the trained model and speed up the fine-tuning phase. Experimental results show that, our ODN can effectively detect and recognize new categories with little human intervention. In the future work, we will further improve the steadiness of our network, and explore its application in a wider range of applications such as abnormal event detection.

\vspace{1em}
\noindent \textbf{Acknowledgement. }This work is partially supported by grants from the National Key R\&D Program of China under grant 2017YFB1002401, the National Natural Science Foundation of China under contract No. U1611461, No. 61390515, No. 61425025, No. 61471042 and No. 61650202.

\small
\bibliographystyle{IEEEbib}
\bibliography{camera-ready_icme2018}

\end{document}